\documentclass[letterpaper, 10 pt, conference]{ieeeconf}
\IEEEoverridecommandlockouts
\usepackage{cite}
\usepackage{amsmath,amssymb,amsfonts}
\usepackage{algorithmic}
\usepackage{graphicx}
\usepackage{textcomp}
\usepackage{xcolor}
\usepackage{svg}
\usepackage{multirow}
\usepackage{algorithm}
\usepackage{bm}
\usepackage{booktabs}
\usepackage[font=scriptsize]{caption} 
\usepackage{subcaption}
\usepackage{todonotes}

\usepackage{lipsum}
\usepackage{float}

\title{\LARGE \bfseries
Population-Coded Spiking Neural Networks for High-Dimensional Robotic Control
}

\author{
\textbf{Kanishkha Jaisankar}$^{1}$ \quad
\textbf{Xiaoyang Jiang}$^{1}$ \quad
\textbf{Feifan Liao}$^{1}$ \quad
\textbf{Jeethu Sreenivas Amuthan}$^{1}$\\
$^{1}$Center for Data Science, New York University, New York, USA\\
\texttt{\{kj2675, xj2366, fl2656, ja5163\}@nyu.edu}
}

\begin{document}

\twocolumn[{
\renewcommand\twocolumn[1][]{#1}
\maketitle
\begin{center}
    \captionsetup{type=figure}
    \includegraphics[width=2.0\columnwidth]{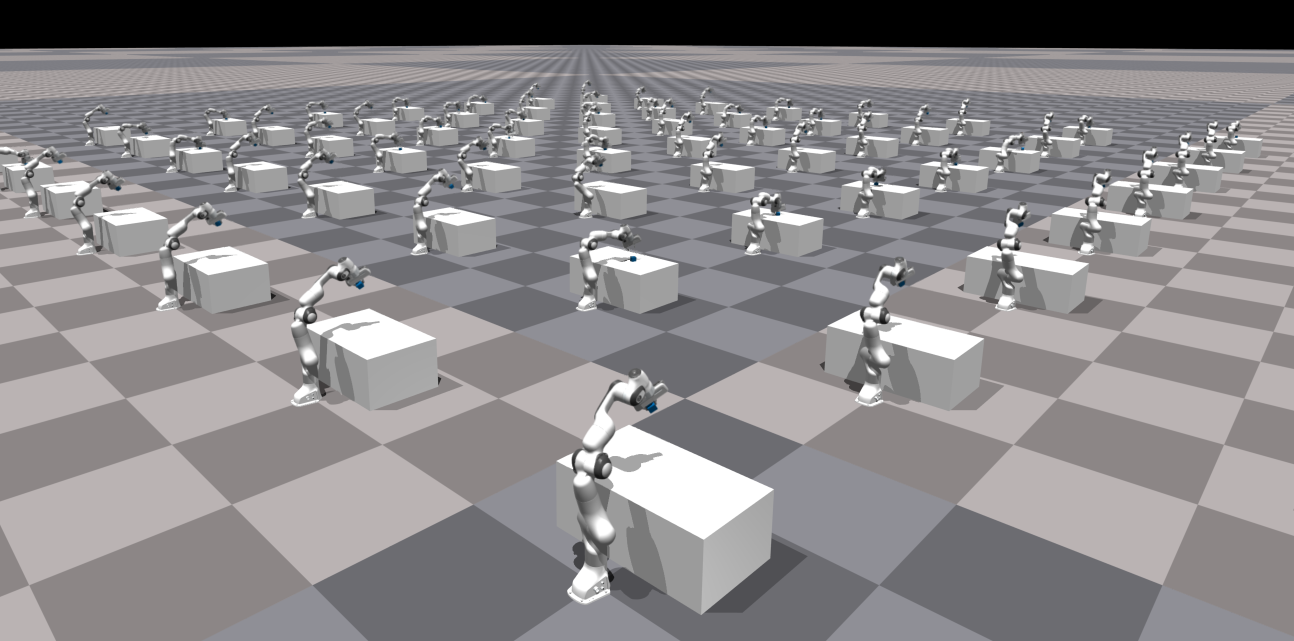}
    \captionof{figure}{Fully spike-based control on robotic arms (Franka) in Isaac Gym}
    \label{fig1}
\end{center}
}]

% \blfootnote{$^{1}$The authors are with Center for Data Science, New York University, New York, USA (\{kj2675, xj2366,  fl2656, ja5163\}@nyu.edu)}

% \maketitle
\thispagestyle{empty}
\pagestyle{empty}

\vspace{-0.37cm}
\begin{abstract}
In recent years, advancements in robotic systems have highlighted the challenges of achieving energy-efficient and high-performance motor control, particularly in high-dimensional continuous control tasks.  This paper introduces a novel framework combining population-coded Spiking Neural Networks (SNNs) and Deep Reinforcement Learning (DRL) to address these challenges. By leveraging the asynchronous, event-driven computation of SNNs and the robust policy optimization capabilities of DRL, our approach achieves a balance between energy efficiency and computational effectiveness. Central to this framework is the Population-coded Spiking Actor Network (PopSAN), which encodes high-dimensional observations into neuronal population activities, enabling optimal policy learning through gradient-based updates. The proposed method is evaluated using the Isaac Gym platform on benchmarks like PixMC featuring complex motor control tasks. Experimental results demonstrate significant improvements in energy efficiency, reduced latency, and robust performance in continuous action spaces, positioning this approach as a promising direction for resource-constrained robotics applications.
\end{abstract}

\section{Introduction}
% The increasing adoption of mobile robots, which are equipped with continuous high-dimensional observations and action space, to tackle a wide range of intricate tasks in real-world situations emphasizes the critical importance of exceptional control algorithms. Currently, the limited onboard energy resources of most robots pose a significant challenge, as this constraint hinders their ability to operate continuously and cost-effectively. Consequently, there is an immediate demand for developing energy-efficient solutions for the seamless control of these autonomous machines.
Over the past decade, robots have been increasingly deployed to handle various complex tasks in real-world scenarios, often characterized by continuous high-dimensional observation and action spaces. With the advancement of neural networks, deep learning technologies have injected new vitality into the control algorithms of these robots. However, most robots still face the challenge of limited onboard resources, which hinders their ability to perform tasks continuously and cost-effectively in certain specific scenarios. Therefore, the development of an energy-efficient and high-performance solution has become an urgent necessity.
% Deep reinforcement learning (DRL) employs deep neural networks (DNNs) as potent function approximators for learning optimal control strategies for intricate tasks \cite{ha2018automated,zhu2017target}, through directly mapping the original state space to the action space \cite{duan2016benchmarking,lillicrap2015continuous}. Nonetheless, the remarkable performance of DRL frequently comes at the expense of substantial energy consumption and slower execution speeds, making them unsuitable for various applications. Additionally, the execution speed of control strategies employing DNNs tends to be slower in comparison to the operational speed of the motion units. This discrepancy often results in a step-like behavior in the control signals, causing negative impacts on the performance of the system. Moreover, the reliance of current Deep Neural Networks (DNNs) on high-power GPUs for computation significantly impacts the structural design of robots, necessitating the inclusion of substantial active cooling systems.
Currently, many tasks adopt Deep Reinforcement Learning (DRL) methods, which leverage Deep Neural Networks (DNNs) as powerful function approximators to directly map the original state space to the action space \cite{duan2016benchmarking,lillicrap2015continuous}, thereby learning optimal control strategies for complex tasks \cite{ha2018automated,zhu2017target}. However, the remarkable performance of DRL often comes at the cost of high energy consumption and slower execution speeds, rendering it unsuitable for many applications. Furthermore, recent works have increasingly incorporated Multimodal Large Language Models (MLLMs)\cite{liu2024self, wang2024lami} or their subsets to achieve a more comprehensive understanding of observation representations, which undoubtedly exacerbates energy consumption and execution speed issues.

Spiking neural networks (SNNs), also referred to as third-generation neural networks, present a promising alternative for energy-efficient and high-speed deep networks. These emerging SNNs operate based on the principles of neuromorphic computing, wherein the integration of memory and computation is seamless, and neurons engage in asynchronous, event-based communication and computation \cite{davies2018loihi}. The biological plausibility, the significant increase in energy efficiency (particularly when deployed on neuromorphic chips \cite{roy2019towards}), high-speed processing and real-time capability for high-dimensional data (especially from asynchronous sensors like event-based cameras \cite{gallego2020event}) contribute to the advantages that SNNs possess over ANNs in specific applications. These advantages render the utilization of SNNs not only feasible but also highly advantageous in lieu of ANNs for performing effective calculations. A mounting body of research illustrates that SNNs can function as energy-efficient and high-speed solutions for effectively managing robot control in scenarios where there are limitations on onboard energy resources \cite{tang2019spiking,taunyazov2020event,michaelis2020robust}. To address the limitations of SNNs in tackling high-dimensional control problems, a natural approach involves combining the energy efficiency of SNNs with the optimality of DRL, which has proven effective in various control tasks \cite{mnih2015human}. Due to the role of rewards as training guides in reinforcement learning (RL), some studies utilize a three-factor learning rule \cite{fremaux2013reinforcement} to implement reward learning. Although these rules exhibit strong performance in low-dimensional tasks, they often struggle to handle complex problems, and the optimization process becomes challenging in the absence of a global loss function \cite{legenstein2005can}. Recently, \cite{rosenfeld2019learning} proposed a strategy gradient-based algorithm to train an SNN for learning random strategies. However, this algorithm is designed for discrete action spaces, and its practical applications are somewhat limited when tackling high-dimensional continuous control problems.

The recent conceptualization of the brain's topology and computational principles has ignited advancements in SNNs, exhibiting both human-like behavior\cite{balachandar2020spiking} and superior performance\cite{kreiser2020chip}. A pivotal attribute associated with effective computation in the brain is the employment of neuronal populations for encoding and representing information, encompassing the entire spectrum from sensory input to output signals. In this scenario, each neuron within a population has a receptive field that captures a specific segment of the encoded signal\cite{georgopoulos1986neuronal}. Notably, initial investigations into this group coding scheme have shown its enhanced capability to represent stimuli\cite{tkavcik2010optimal}, contributing to recent triumphs in training SNNs for complex, high-dimensional supervised learning tasks\cite{bellec2018long,pan2019neural}. The efficacy of population coding presents a promising pathway for the advancement of efficient SNNs that leverage population coding. These networks possess the potential to acquire optimal solutions for complex high-dimensional continuous control tasks, paving the way for significant progress in this field.

The main contributions of this paper are summarized as follows:
\begin{itemize}
\item We propose a novel hybrid framework that integrates Spiking Neural Networks (SNNs) with Deep Reinforcement Learning (DRL) to address energy efficiency and high-dimensional continuous control tasks.
\item Our approach leverages population coding to enhance the representational capacity of SNNs, enabling efficient learning and control in high-dimensional robotic applications.
\item We demonstrate the effectiveness of our framework through extensive experiments on complex robotic control tasks, showcasing significant improvements in energy efficiency, execution speed, and control performance.
\end{itemize}

\section{Related work}

% \subsection{RL related(RL + MLLM...)}
\subsection{Self-Supervised Learning (SSL)}

Recent advancements in self-supervised learning (SSL) have significantly impacted motor control and vision-based tasks by enabling robust feature extraction from large-scale unlabeled datasets. Techniques such as Masked Autoencoders (MAE) \cite{he2022masked} have demonstrated remarkable success in pre-training visual representations by reconstructing masked image patches, leading to scalable solutions for reinforcement learning (RL) tasks. Contrastive learning methods, like Contrastive Unsupervised Representations for Reinforcement Learning (CURL)\cite{laskin2020curl}, have also shown that augmenting the training data with auxiliary objectives can improve representation quality and sample efficiency in RL. Furthermore, frameworks like Dense Object Nets\cite{florence2018dense} have leveraged SSL to learn dense visual object descriptors, which are crucial for robotic manipulation tasks involving complex object interactions.

\subsection{Spiking Neural Networks (SNN)}

A major challenge in robotics lies in efficiently processing high-dimensional sensory data. While artificial neural networks (ANNs) have been dominant, Spiking Neural Networks (SNNs) are gaining traction for their energy efficiency and temporal processing capabilities. SNNs, which mimic the spiking behavior of biological neurons, have been applied in real-time robotics tasks such as object detection and tracking\cite{panda2020toward}, and as highlighted by Pfeiffer and Pfeil\cite{pfeiffer2018deep}, they show promise for neuromorphic hardware integration. However, their potential for integration into complex motor control tasks remains underexplored. \cite{tavanaei2019deep} provide a comprehensive review of SNNs in neuromorphic computing, suggesting their suitability for energy-efficient robotics applications.

\subsection{vision-language models (VLMs)}

The rise of vision language models (VLMs) and multimodal learning frameworks has introduced new paradigms in robotics. Models such as CLIP \cite{radford2021learning} and FLAVA \cite{singh2022flava} leverage large-scale pre-training across visual and textual modalities, enabling semantic understanding and contextual reasoning. For example, CLIP has demonstrated superior generalization in various vision tasks by aligning visual features with text embeddings, which can be critical for robotics applications where high-level semantic reasoning is required for task execution. Furthermore, pre-trained multimodal models such as Align before Fuse \cite{li2021align} have shown promising results in improving the interpretability of robotic perception systems.

\subsection{Reinforcement learning (RL) for Motor Control}

Reinforcement learning (RL) remains a cornerstone for learning motor control policies, particularly in high-dimensional environments. Proximal Policy Optimization (PPO) \cite{schulman2017proximal} and other model-free RL methods have become popular for learning complex control tasks. Studies such as OpenAI’s dexterous manipulation research \cite{kumar2016learning} and scalable RL for vision-based robotic manipulation \cite{kalashnikov2018scalable} have demonstrated the effectiveness of RL when combined with robust visual representations. However, these methods often suffer from high sample complexity and limited generalization. Auxiliary objectives and pre-trained encoders, such as those proposed in this study, are potential solutions to these challenges.

\subsection{Benchmarks for Motor Control}

Robust benchmarks play a critical role in evaluating motor control and vision models. Existing platforms, including the DeepMind Control Suite \cite{tassa2018deepmind}, RLBench \cite{james2020rlbench} and Meta-World \cite{yu2020meta} provide standardized environments for assessing RL algorithms and visual encoders. PixMC \cite{xiao2022masked} , for example, offers diverse tasks, dense rewards, and high-resolution pixel observations, which are particularly valuable for evaluating SSL-based visual encoders like MAE. These benchmarks have significantly accelerated research in generalization and robustness of visual representations.

\subsection{Multimodal Large Langugae Models in Robotics}

Incorporating multimodal large language models (LLMs) into robotic systems is an exciting frontier. Models such as Flamingo \cite{alayrac2022flamingo} have showcased the potential of integrating text and vision to enhance context understanding in real-time systems. Similarly, studies on energy-efficient multimodal encoders \cite{he2022masked} highlight the feasibility of deploying such systems on hardware-constrained robotics platforms. Furthermore, \cite{driess2023palm} introduced PaLM-E: An Embodied Multimodal Language Model, which integrates vision and language in an embodied context, enabling robots to perform tasks that require both physical interaction and semantic understanding. The model demonstrates generalization across tasks and modalities, making it a strong choice for robotics applications.

\subsection{Representation learning strategies for RL}
Moreover, prior research has explored various representation learning strategies for RL. CURL \cite{laskin2020curl}, RAD \cite{laskin2020reinforcement}, and MoCo \cite{chen2020simple} have proposed ways to improve sample efficiency by integrating augmentation and contrastive objectives into the training process. These techniques have paved the way for integrating sophisticated encoders, such as MAE and Vision Transformers \cite{he2022masked}, into the motor control pipeline, enabling efficient learning from high-dimensional pixel inputs. 

Despite these advancements, a unified approach combining SSL, SNNs, and VLMs for motor control remains largely unexplored. Our study addresses this gap by replacing traditional ANNs with SNNs in motor control tasks and exploring the potential of multimodal encoders to enhance both perceptual and control capabilities in robotics. By building on benchmarks like PixMC and techniques like CURL, we aim to push the boundaries of energy-efficient, generalizable robotic systems.

\section{Method}

To evaluate the effectiveness of our algorithm, we conducted training and testing using Isaac Gym \cite{makoviychuk2021isaac}. Isaac Gym is a high-performance simulation platform specifically designed for robotics applications. It provides realistic physics simulations, tailored support for all kinds of robots, seamless integration with NVIDIA technologies, and extensive customization options. Additionally, it benefits from a robust and active community, making it an ideal environment for developing and testing robotics algorithms.

\begin{figure*}[tbp]
\vspace{0.13cm}
\setlength{\abovecaptionskip}{0cm}
\setlength{\belowcaptionskip}{-0.3cm}
\centerline{\includegraphics[width=2.0\columnwidth]{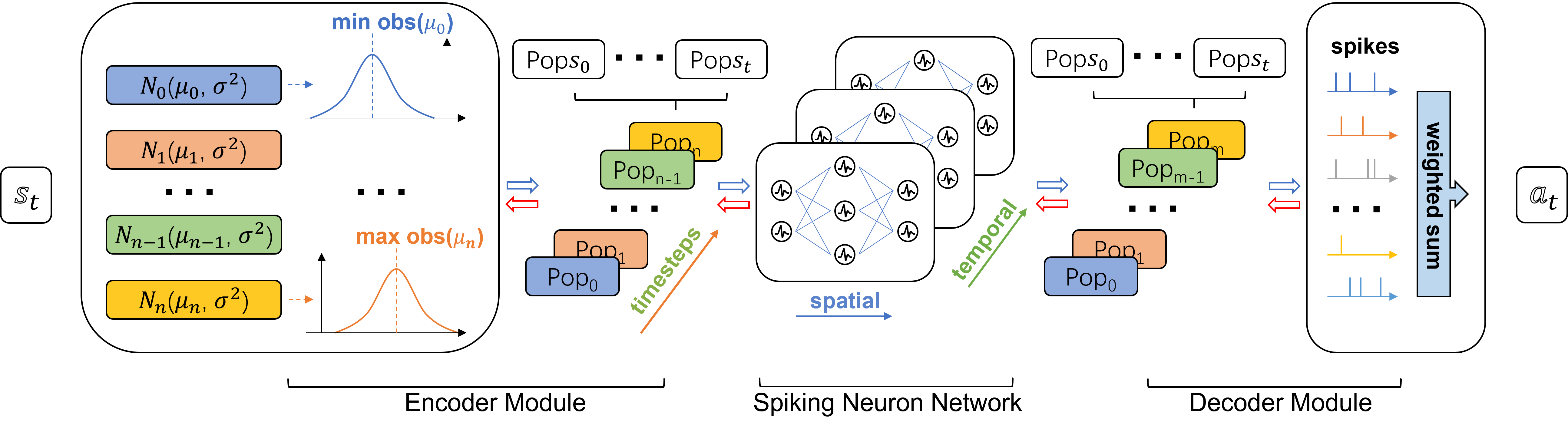}}
\caption{The observations are initially encoded by the encoder as $n$ independent distributions that are uniformly distributed over the observation range. After encoding, the population processes the distributions, resulting in spike generation. The neurons in the input populations encode each observation dimension and drive a multi-layered, fully connected SNN. During forward timesteps in PopSAN, the activities of each output population are decoded to determine the corresponding action dimension. This implies that the neural network receives observations, processes them using the SNN, and decodes the resulting activities to determine the appropriate action for the specific situation.}
\label{fig2}
\end{figure*}

\subsection{SNN policy}
We employ a population-coded spiking actor-network (PopSAN)\cite{tang2021deep} that is trained in tandem with a deep critic network using the DRL algorithms. During training, the PopSAN generated an action $\alpha$ $\in$ $\mathbb{R}^{N}$ for a given observation, $s$, and the deep critic network predicted the associated state value $V$($s$) or action-value $Q$($s$, $\alpha$), which in turn optimized the PopSAN, in accordance with a chosen DRL method (Fig. \ref{fig2}). Within the PopSAN architecture, the encoder module is responsible for encoding individual dimensions of the observation by mapping them to the activity of distinct neuron populations. During forward propagation, the input populations activate a multi-layer fully-connected SNN. The SNN then produces activity patterns within the output populations. At the end of each set of $T$ timesteps, these patterns of activity are decoded in order to ascertain the associated action dimensions. (as outlined in Algorithm \ref{alg:I}).

The current-based leaky-integrate-and-fire (LIF) model of a spiking neuron is employed in constructing the SNN. This model is utilized in constructing the SNN architecture. The dynamics of the LIF neurons are governed by a two-step model, as elaborated in Algorithm \ref{alg:I}: i) the integration of presynaptic spikes $o$ into current $c$; and ii) the integration of current $c$ into membrane voltage $v$; $d_c$ and $d_v$ represent the current and voltage decay factors, respectively. In this particular implementation, a neuron fires a spike when its membrane potential surpasses a predetermined threshold. The hard-reset model was implemented, in which the membrane potential is promptly reset to the resting potential following a spike. The resultant spikes are transmitted to the post-synaptic neurons during the same inference timestep, under the assumption of zero propagation delay. This approach facilitates efficient and synchronized information transmission within the SNN.

\begin{algorithm}[tbp] 
\caption{Forward propagation through PopSAN} 
\label{alg:I}
\begin{algorithmic}[] 

\STATE Randomly initialize weight matrices $\bm{W}$ and biases $\bm{b}$ for each SNN layer;\\
\STATE Initialize encoding means $\bm{\mu}$ and standard deviations $\bm{\sigma}$ for all input populations;\\
\STATE Randomly initialize decoding weight vectors $\bm{W}_d$ and bias $b_d$ for each action dimension;\\
\STATE N-dimensional observation, $\bm{s}$;\\
\STATE Spikes from input populations generated by encoder module: $\bm{X}$ = Encoder($\bm{s}$, $\bm{\mu}$, $\bm{\sigma}$);\\

\FOR{$t=1,...,T$}
\STATE Spikes at timestep t: $\bm{o}^{(t)(0)} = \bm{X}^{(t)}$;
\FOR{$k=1,...,K$}
\STATE Update LIF neurons in layer $k$ at timestep $t$ based on spikes from layer $k - 1$:
\STATE $\bm{c}^{(t)(k)} = d_c\cdot\bm{c}^{(t-1)(k)} + \bm{W}^{(k)}\bm{o}^{(t)(k-1)} + \bm{b}^{(k)}$;
\STATE $\bm{v}^{(t)(k)} = d_c\cdot\bm{v}^{(t-1)(k)}\cdot(1 - \bm{o}^{(t-1)(k)}) + \bm{c}^{(t)(k)}$;
\STATE $\bm{o}^{(t)(k)} = Threshold(\bm{v}^{(t)(k)})$;
\ENDFOR
\ENDFOR
\STATE $M$-dimensional action $\bm{a}$ generated by decoder module:
\STATE Sum up the spikes of output populations: $\bm{sc} = \sum_{t=1}^T\bm{o}^{(t)(K)}$;
\FOR{$i=1,...,M$}
\STATE Compute firing rates of the $i^{th}$ output population:
$\bm{fr}^{(i)} = \bm{sc}^{(i)}/T$;
\STATE Compute $i^{th}$ dimension of action: $\alpha^i = {\bm{W}_d}^{(i)}\cdot\bm{fr}^{(i)} + {b_d}^{(i)}$;
\ENDFOR 
\end{algorithmic}
\end{algorithm}

% \subsection{SNN Multimodal encoder}
% the SNN.

\subsection{Training}
In our study, we used gradient descent to update the PopSAN parameters, with the specific loss function. To train PopSAN parameters, we use the gradient of the loss with respect to the computed action, denoted as $\nabla_a$$L$. The parameters for each output population $i, i \in 1, ..., M$ are updated independently as follows:
\begin{equation}
\setlength\abovedisplayskip{4pt}
\setlength\belowdisplayskip{4pt}
\begin{aligned}
\nabla_{{\bm{W}_d}^{(i)}}L = \nabla_{\alpha_i}L\cdot{\bm{W}_d}^{(i)}\cdot\bm{fr}^{(i)}, \nabla_{{b_d}^{(i)}}L = \nabla_{\alpha_i}L\cdot{\bm{W}_d}^{(i)}
\label{eq4}
\end{aligned}
\end{equation}
The SNN parameters are updated using extended spatiotemporal backpropagation as introduced in \cite{tang2020reinforcement}. We utilized the rectangular function $z(v)$, as defined in \cite{wu2018spatio}, to estimate the spike's gradient. The gradients of the loss with respect to the parameters of the SNN for each layer $k$ are computed by aggregating the gradients backpropagated from all timesteps:
\begin{equation}
\setlength\abovedisplayskip{3pt}
\setlength\belowdisplayskip{4pt}
\begin{aligned}
\nabla_{{\bm{W}}^{(k)}}L=\sum_{t=1}^T\bm{o}^{(t)(k-1)}\cdot\nabla_{{\bm{c}}^{(t)(k)}}L, \nabla_{{\bm{b}}^{(k)}}L=\sum_{t=1}^T\nabla_{{\bm{c}}^{(t)(k)}}L 
\label{eq5}
\end{aligned}
\end{equation}
Lastly, we updated the parameters independently for each input population $i, i \in 1, ..., N$ as follows: 
\begin{equation}
\setlength\abovedisplayskip{4pt}
\setlength\belowdisplayskip{4pt}
\begin{aligned}
\nabla_{{\bm{\mu}}^{(i)}}L=&\sum_{t=1}^T\nabla_{{\bm{o}_i}^{(t)(o)}}L\cdot\bm{A_E}^{(i)}\cdot\frac{s_i - \bm{\mu}^{(i)}}{\bm{\sigma^{(i)^2}}}, \\
\nabla_{{\bm{\sigma}}^{(i)}}L=&\sum_{t=1}^T\nabla_{{\bm{o}_i}^{(t)(o)}}L\cdot\bm{A_E}^{(i)}\cdot\frac{{(s_i - \bm{\mu}^{(i)})}^2}{\bm{\sigma^{(i)^3}}}
\label{eq6}
\end{aligned}
\end{equation}

\section{Experiments}

The objectives of our experiments were to validate the feasibility of employing Population-Coded Spiking Neural Networks (PopSAN) in high-dimensional, continuous control tasks, evaluate their benefits over traditional Artificial Neural Networks (ANNs) in terms of energy efficiency, execution speed, and performance consistency, and benchmark our proposed hybrid SNN-DRL framework against existing state-of-the-art methods with respect to energy consumption, latency, and control accuracy.

\subsection{Simulation Setup}
The simulations were conducted using the NVIDIA Isaac Gym platform, a high-performance GPU-based physics simulator designed for robotics applications. We using an Ubuntu 20.04 LTS system  with an NVIDIA T4 GPU. The code has been tested with PyTorch 1.10, CUDA 11.3, and cuDNN 8.2. For the experiments, we selected two robots, Franka and KUKA, representing robotic arms, to test the scalability of our approach across varying degrees of freedom (DoF) and control complexities.

The environments were constructed using the PixMC benchmark, which includes a suite of tasks specifically designed for motor control from pixel-based observations. Our experiments focused exclusively on the Pick action, one of the 4 tasks available in PixMC, to ensure controlled and reproducible testing conditions.

The control policies were trained using 20,000 iterations for ANN-based methods and 40,000 iterations for SNN-based methods. During training, sensory inputs from the robots, including joint positions and velocities, were fed into the policy networks to produce continuous control actions. To ensure comprehensive monitoring and analysis, we employed Weights \& Biases (wandb) to collect and log key metrics throughout the training process.  This setup allowed for rigorous evaluation of both the performance and energy efficiency of our proposed hybrid SNN-DRL framework under realistic robotic scenarios.

\subsection{Training Comparison}

\subsubsection{Mean Success Rate}
\begin{figure}[ht]
    \centering
    % Side-by-side plots for SNN and ANN
    \begin{subfigure}{0.45\columnwidth}
        \centering
        \includegraphics[width=\textwidth]{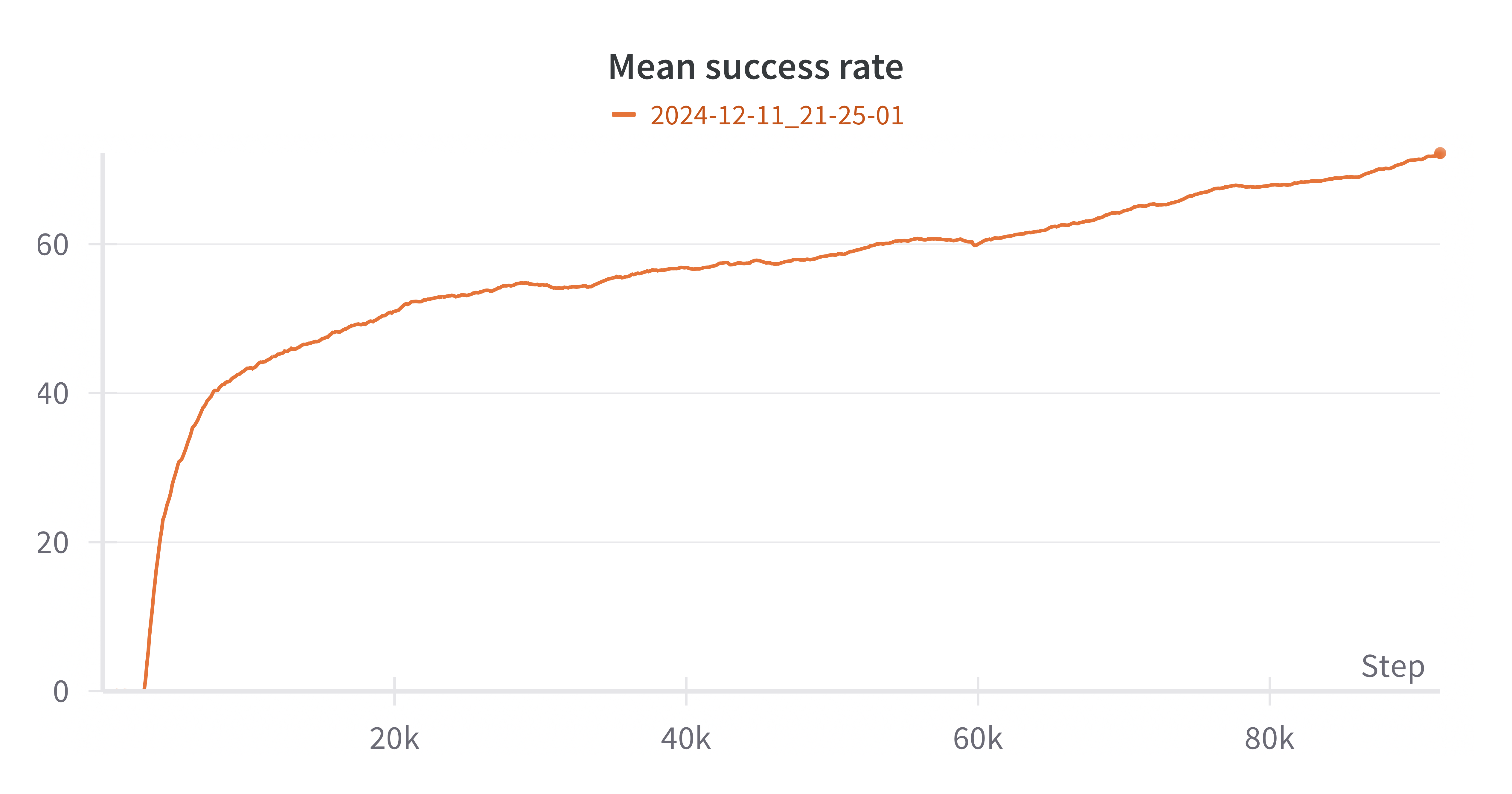}
        \caption{SNN Mean Success Rate}
        \label{fig:mean_success_rate_snn}
    \end{subfigure}
    \hspace{0.05\columnwidth} % Spacing between the plots
    \begin{subfigure}{0.45\columnwidth}
        \centering
        \includegraphics[width=\textwidth]{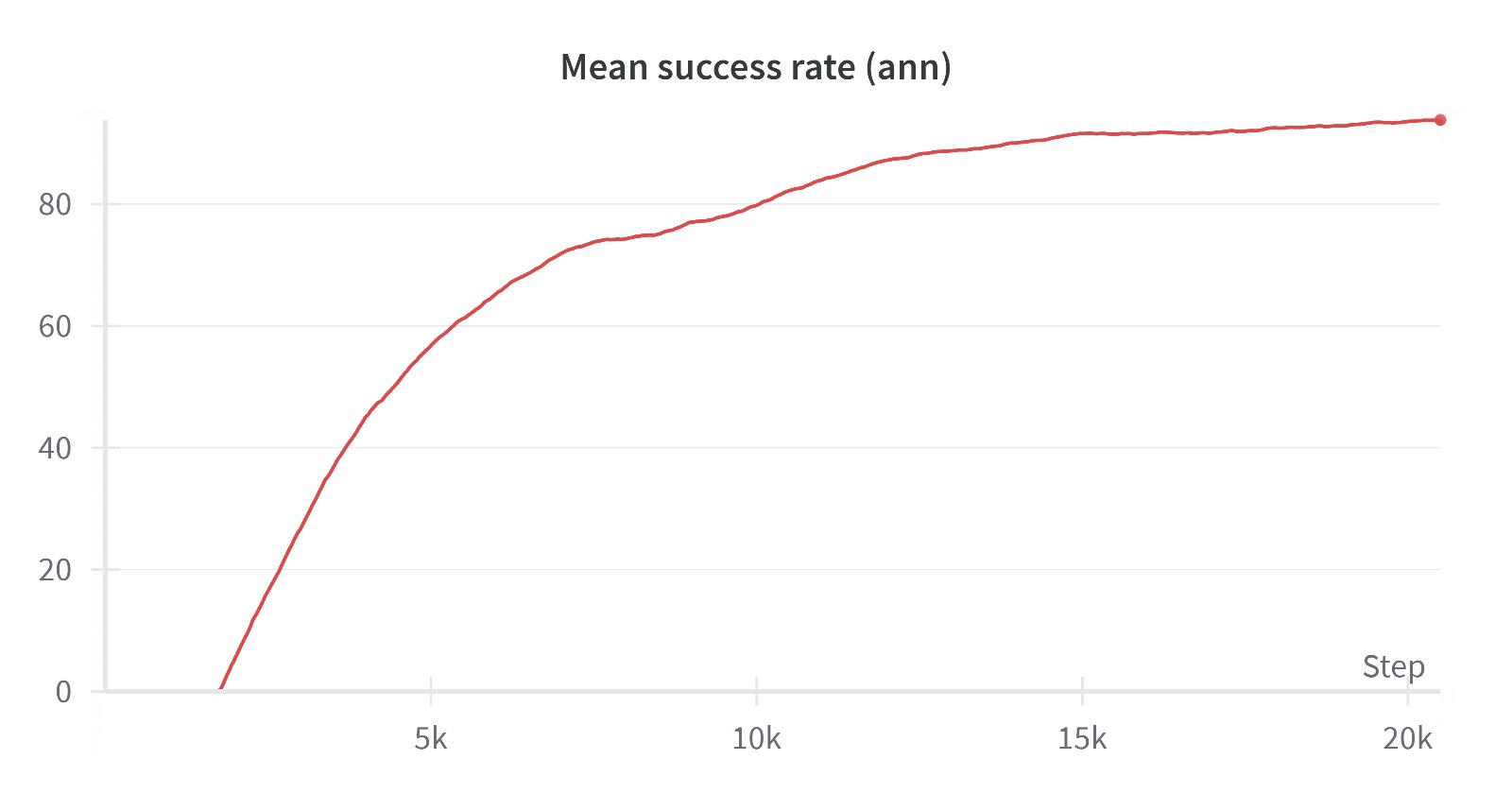}
        \caption{ANN Mean Success Rate}
        \label{fig:mean_success_rate_ann}
    \end{subfigure}
    \caption{Comparison of Mean Success Rate for SNN and ANN during training.}
    \label{fig:mean_success_rate_comparison}
\end{figure}

The SNN-based approach shows a slow and irregular progression in the mean success rate due to the event-driven, temporally asynchronous architecture of spiking neural networks. This architecture complicates gradient calculations and introduces variability in learning, resulting in slower convergence. In contrast, ANNs demonstrate a much smoother and faster increase in success rate, achieving high success levels early in training. Despite the slower learning process, SNNs achieve comparable success rates after extended training, with the added benefit of energy-efficient inference.

\subsubsection{Mean Episode Length}
\begin{figure}[ht]
    \centering
    % Side-by-side plots for SNN and ANN
    \begin{subfigure}{0.45\columnwidth}
        \centering
        \includegraphics[width=\textwidth]{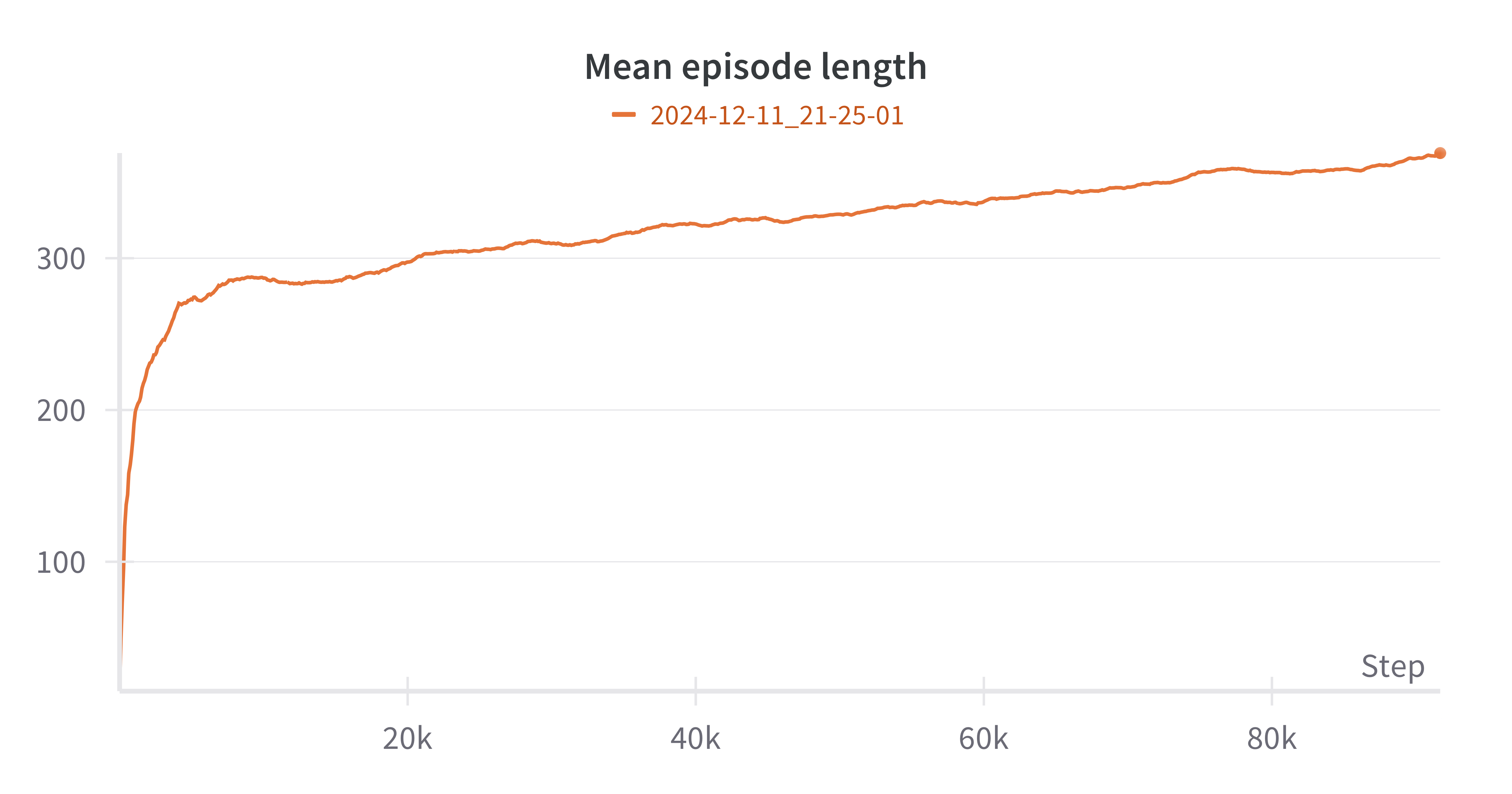}
        \caption{SNN Mean Episode Length}
        \label{fig:mean_episode_length_snn}
    \end{subfigure}
    \hspace{0.05\columnwidth} % Spacing between the plots
    \begin{subfigure}{0.45\columnwidth}
        \centering
        \includegraphics[width=\textwidth]{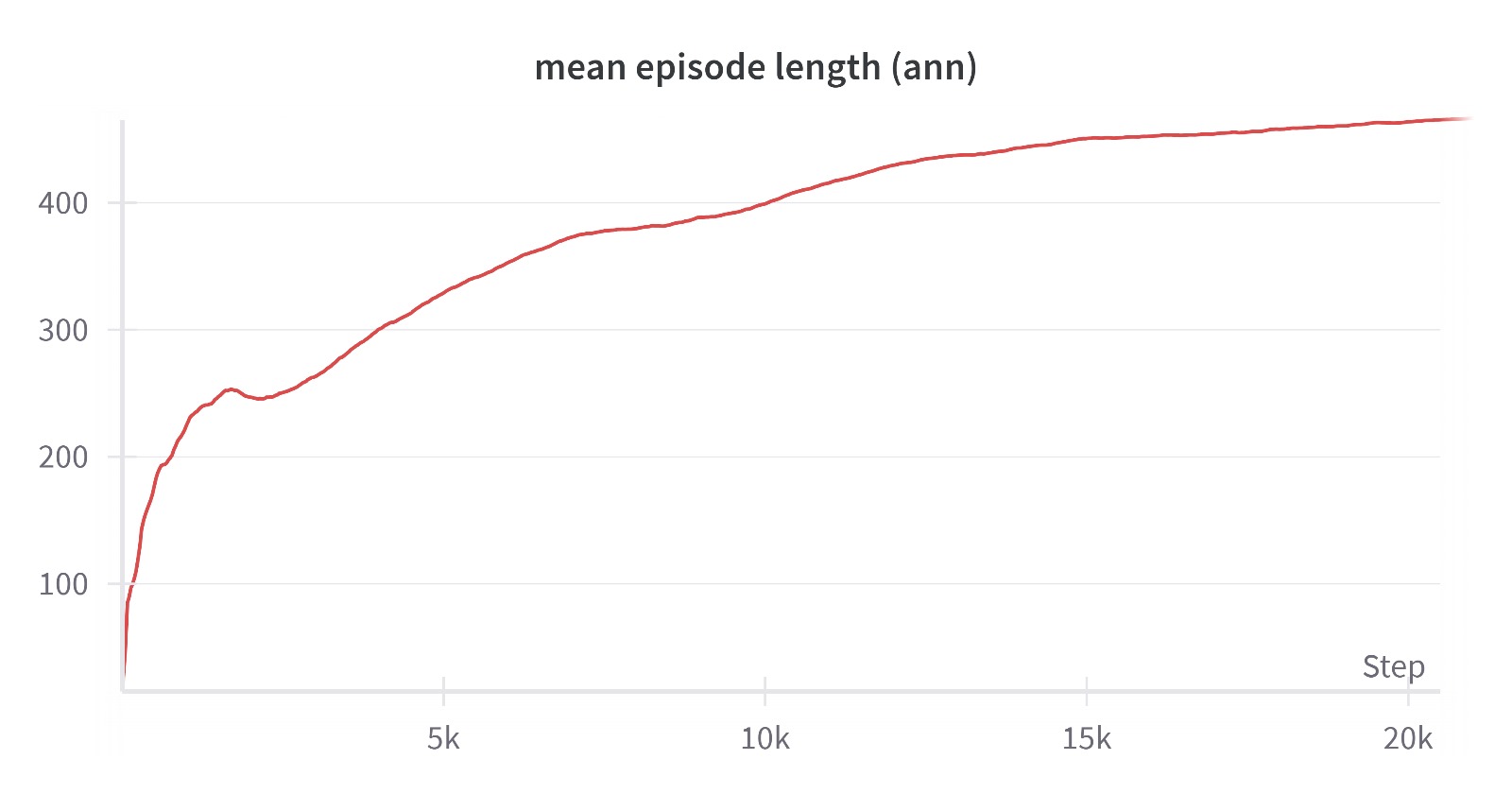}
        \caption{ANN Mean Episode Length}
        \label{fig:mean_episode_length_ann}
    \end{subfigure}
    \caption{Comparison of Mean Episode Length for SNN and ANN during training.}
    \label{fig:mean_episode_length_comparison}
\end{figure}

For the mean episode length, SNNs exhibit significant fluctuations during training, reflecting the temporal dependency of their spiking mechanisms and the delayed propagation of information. This leads to a slower stabilization of policies compared to ANNs, which leverage deterministic gradients and continuous-valued activations for faster policy learning. Despite this, SNNs can maintain stable and efficient operation during inference, showcasing their adaptability after training is complete.

\subsubsection{Mean Reward}
\begin{figure}[ht]
    \centering
    % Side-by-side plots for SNN and ANN
    \begin{subfigure}{0.45\columnwidth}
        \centering
        \includegraphics[width=\textwidth]{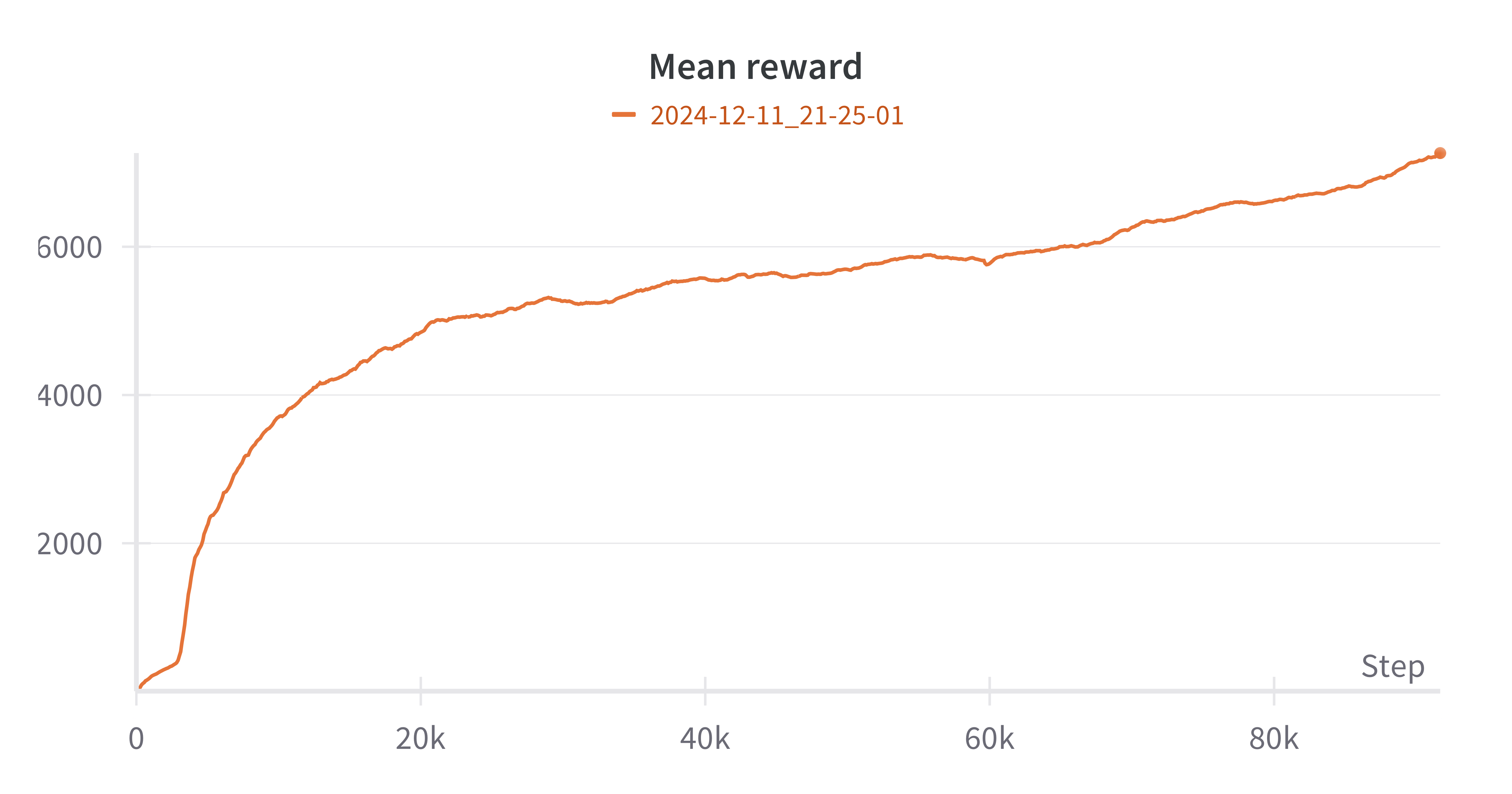}
        \caption{SNN Mean Reward}
        \label{fig:mean_reward_snn}
    \end{subfigure}
    \hspace{0.05\columnwidth} % Spacing between the plots
    \begin{subfigure}{0.45\columnwidth}
        \centering
        \includegraphics[width=\textwidth]{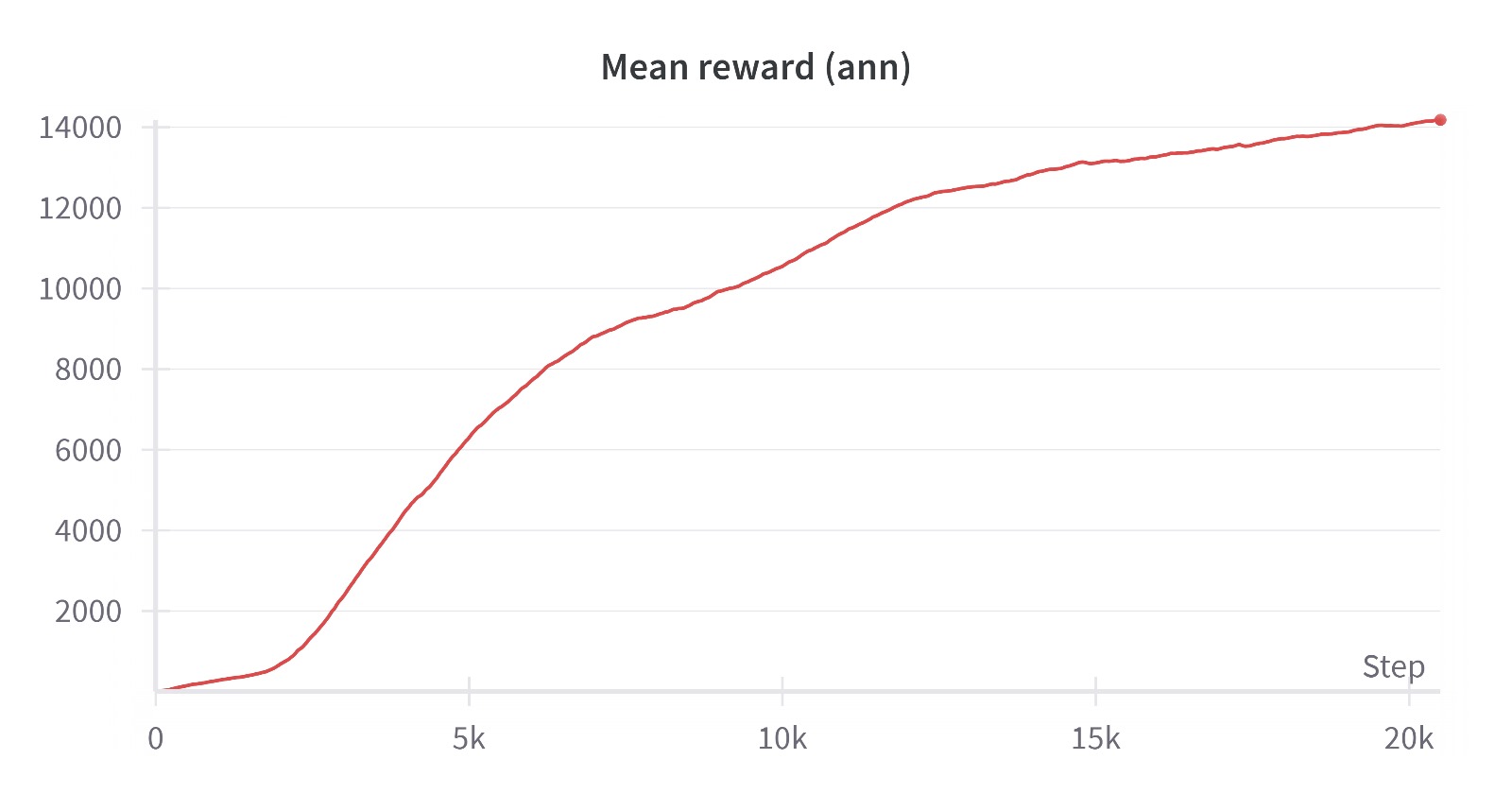}
        \caption{ANN Mean Reward}
        \label{fig:mean_reward_ann}
    \end{subfigure}
    \caption{Comparison of Mean Reward for SNN and ANN during training.}
    \label{fig:mean_reward_comparison}
\end{figure}

The mean reward metric further highlights the slow convergence of SNNs, attributed to the difficulty in encoding reward signals within the temporal dynamics of spiking architectures. SNNs rely on surrogate gradients, which introduce complexity and noise in optimization, leading to a delayed improvement in rewards. On the other hand, ANNs quickly and consistently achieve higher rewards due to their straightforward gradient-based optimization. Nevertheless, once trained, SNNs reach similar reward levels while offering superior energy efficiency and faster inference.

\subsubsection{Conclusion}
The comparison of training metrics—Mean Success Rate, Mean Episode Length, and Mean Reward—demonstrates the inherent challenges in training spiking neural networks. Their unique event-driven architecture and reliance on temporal dynamics result in slower and noisier convergence compared to artificial neural networks. However, SNNs excel in inference efficiency and energy usage, making them a compelling choice for applications requiring long-term efficiency and low-power operation. While ANNs achieve faster training convergence, SNNs prove to be competitive once properly trained, providing substantial advantages in real-world scenarios.

\subsection{Performances of SNNs control}

\begin{figure}[ht]
    \centering
    % Replace 'figure3.png' with the actual filename/path of your figure
    \includegraphics[width=1\columnwidth]{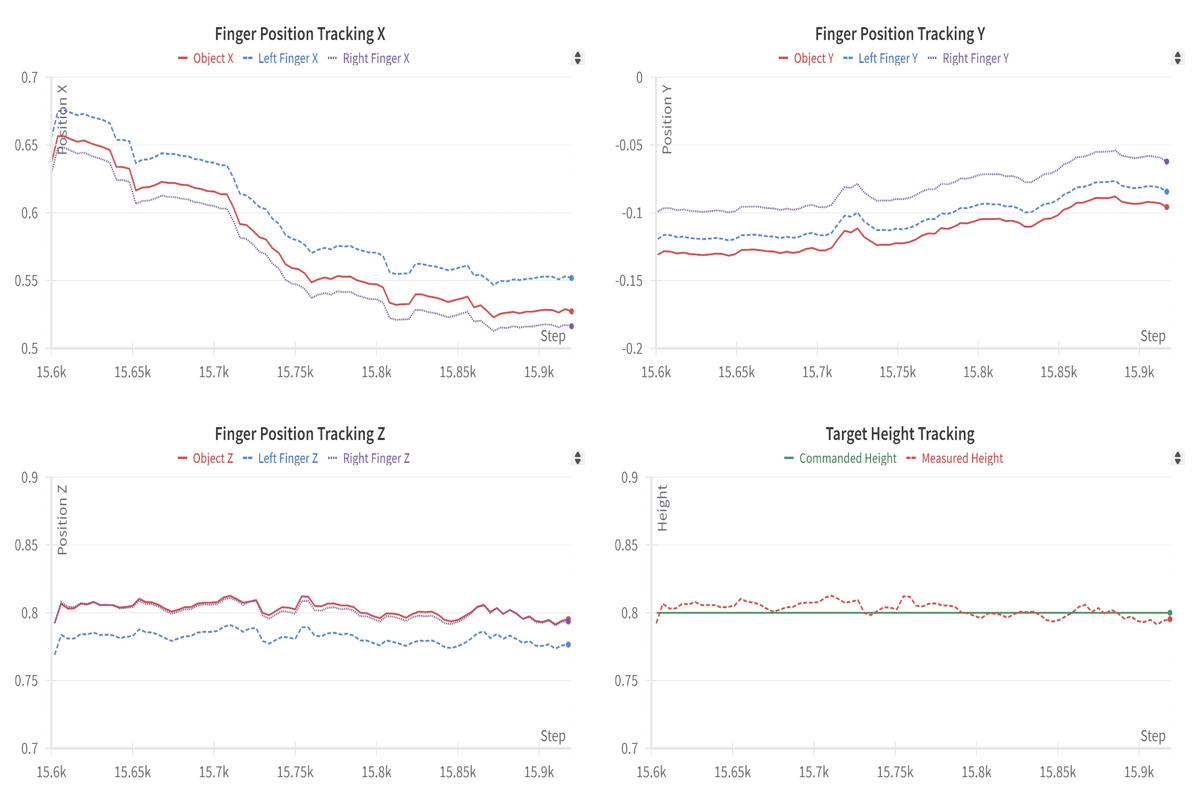}
    \caption{%
        Combined plots of Finger Position Tracking (X, Y, and Z axes) and Target Height Tracking for the Franka pick task. The red line denotes the commanded object trajectory (object position), while the blue and purple lines represent the measured positions of the left and right fingers respectively. The green line in the target height subplot represents the commanded height, and the red dashed line shows the measured height.
    }
    \label{fig:snns_control_performance}
\end{figure}

\subsubsection{Finger Position Tracking (X, Y, Z Axes)}
In the given plots (Figures: Finger Position Tracking X, Finger Position Tracking Y, and Finger Position Tracking Z), the red line represents the commanded object position trajectory—essentially the target path that the Franka robotic arm’s end-effector and fingers aim to follow. The blue (left finger) and purple (right finger) lines represent the measured finger positions as the robot attempts to pick and stabilize the object. Across all three axes (X, Y, and Z), the proximity of these lines indicates how closely the actual finger movements align with the desired object trajectory.

\paragraph{X-Axis Performance:}  
The figure for the X-axis tracking shows that both fingers track the object position with relatively small error margins. Although some minor deviations occur—particularly at certain time steps—overall, the left and right finger positions remain close to the commanded object line. This suggests that the control approach is effective in guiding the fingers along the intended horizontal direction, ensuring that the gripper closes in accurately on the target.

\paragraph{Y-Axis and Z-Axis Performance:}  
In the Y and Z axes, similar trends are observed. While slight offsets may appear, the measured finger positions (blue and purple) generally remain near the commanded trajectory (red). In the vertical (Z) direction, the closeness of the tracking lines is especially important for stable lifts and placements. The plots confirm that even as the manipulator adjusts in height (Z), the fingers manage to remain aligned with the object’s commanded position.

These results are indicative that the underlying control system—incorporating Spiking Neural Networks (SNNs) within a Masked Visual Perturbation framework—can produce finger trajectories that closely mirror the ideal path. This is crucial for tasks involving delicate or precise picks, as the robot must react dynamically to maintain proper grip and avoid losing the object.

\subsubsection{Target Height Tracking}
The figure highlighting Target Height Tracking shows both the commanded vertical position (green line) and the measured height (red dashed line) as the robot attempts to hold the object at a specified elevation. The measured line remains fairly close to the commanded line, suggesting that once the object is picked, the system stabilizes the vertical position effectively. Small oscillations or deviations are visible, but they remain within a narrow band, demonstrating stable height maintenance.

\subsection{Energy Consumption}

As mentioned earlier, one of the primary advantages of our SNN policy is its minimal energy usage. The assessment of energy consumption in a SNN is complex because the floating-point operations (FLOPs) in the initial encoder layer are MAC, whereas all other Conv or FC layers are AC. Building upon prior research\cite{hu2021advancing, kundu2021hire, yin2021accurate, yao2023attention} conducted by SNN, it is assumed that the data utilized for various operations is represented in 32-bit floating-point format in 45nm technology\cite{yin2021accurate}, with $E_{MAC}$ = 4.6pJ and $E_{AC}$ = 0.9pJ. The energy consumption equations simulated for SNN are provided as follows:

{\setlength\abovedisplayskip{-6pt}
\setlength\belowdisplayskip{4pt}
\begin{equation}
\begin{aligned}
E_{model} = & E_{MAC} \cdot FL^1_{SNNConv} + \\ & E_{AC} \cdot (\sum_{n=2}^NFL^n_{SNNConv} + \sum_{m=1}^MFL^m_{SNNFC})
\end{aligned}
\end{equation}}

Experimental results (Table \ref{tab2}) demonstrate that our approach offers a significant energy efficiency improvement compared to the conventional ANN architecture. Specifically, it achieves energy savings of $95.65\%$, $81.49\%$, and $60.34\%$ at T = 1, 2 and 3, respectively.

\begin{table}[tbp]
\setlength\belowdisplayskip{1cm}
\renewcommand{\arraystretch}{1.3}
\caption{Energy Comparison($\times10^{-6}$ mJ)}
\vspace{-0.2cm}
\begin{center}
\begin{tabular}{cccc}
\hline
\specialrule{0em}{0.2pt}{0.1pt}
Method & Actor(T=1) & 
Actor(T=2) & Actor(T=3) \\
\specialrule{0em}{0.1pt}{0.1pt}
\hline
\specialrule{0em}{1pt}{0.1pt}
ANN Model & 85.96 & 85.96 & 85.96 \\
SNN Model(\textbf{ours}) & \boldmath{$3.35$} & \boldmath{$15.36$} & \boldmath{$35.22$} \\
\specialrule{0em}{-1pt}{-1pt}
Energy Saving & 96.10\% & 82.13\% & 59.03\% \\
\hline
\end{tabular}
\label{tab2}
\end{center}
\vspace{-0.6cm}
\end{table}

\section{Conclusion}

This work proposed a hybrid framework that integrates population-coded SNNs with DRL to address the critical challenges of energy efficiency and high-dimensional continuous control in robotics. Using event-driven SNN computation and the robust optimization strategies of DRL, the framework achieved a balance between computational effectiveness and energy efficiency. Comprehensive experiments carried out on the Isaac Gym platform demonstrated the ability of the framework to significantly reduce energy consumption and improve execution speeds while maintaining robust control performance in various robotic tasks.

The current implementation was tested exclusively with the Franka robotic arm, focusing solely on the pick action. Future work will extend this framework by training it with different robots (Kuka, etc.,.) and a variety of actions (Reach, Move, Cabinet, etc.,.)  to evaluate its scalability and robustness. Additionally, we plan to integrate Large Language Models (LLMs) into the robotic control pipeline by designing structured prompts that align with the robot's task objectives and operational context. These prompts will facilitate the decomposition of complex goals into hierarchical, actionable instructions that are dynamically adjusted based on sensory feedback and environmental conditions \cite{ma2024explorllm}. This integration will enable robots to execute LLM-guided actions with precision across diverse tasks such as manipulation, navigation, and collaboration \cite{ding2023task}. We are planning to use Multimodal Large Language Models (MLLMs) encoders to enable the fusion of textual commands, visual inputs, and other modalities into coherent action strategies. Furthermore, Vision-Language Models (VLMs) will play a critical role in enabling advanced visual reasoning capabilities, such as semantic scene understanding, object recognition, and context-aware decision-making\cite{zhao2024agentic, dalal2024plan}. By incorporating these advanced models, we aim to develop robotic systems with a high degree of adaptability, multimodal perception, and energy-efficient operation suited for complex real-world applications.

\bibliographystyle{IEEEtran}
\bibliography{ref}{}

\end{document}